\documentclass[letterpaper]{article} 
\usepackage{aaai25}  
\usepackage{times}  
\usepackage{helvet}  
\usepackage{courier}  
\usepackage[hyphens]{url}  
\usepackage{graphicx} 
\usepackage{natbib}  
\usepackage{caption} 
\usepackage{amsmath}
\usepackage{amssymb}
\usepackage{booktabs} 
\usepackage{multicol}
\usepackage{multirow}
\usepackage{algorithm}
\usepackage{algorithmic}
\usepackage{xcolor}
\usepackage{lipsum}
\usepackage{newfloat}
\usepackage{listings}
\DeclareCaptionStyle{ruled}{labelfont=normalfont,labelsep=colon,strut=off} 
\floatstyle{ruled}
\newfloat{listing}{tb}{lst}{}
\floatname{listing}{Listing}
\title{Trusted Unified Feature-Neighborhood Dynamics for Multi-View Classification}
\author{
    Haojian Huang\textsuperscript{\rm 1, \rm 2}, 
    Chuanyu Qin\textsuperscript{\rm 3}, 
    Zhe Liu\textsuperscript{\rm 4}, 
    Kaijing Ma\textsuperscript{\rm 1, \rm 5}, 
    Jin Chen\textsuperscript{\rm 1, \rm 5}, 
    Han Fang\textsuperscript{\rm 1}, 
    Chao Ban\textsuperscript{\rm 1}, 
    Hao Sun\textsuperscript{\rm 1}\thanks{Corresponding author}, 
    Zhongjiang He\textsuperscript{\rm 1}\footnotemark[1]
}
\affiliations{
    \textsuperscript{\rm 1}TeleAI, China\\
    \textsuperscript{\rm 2}The University of Hong Kong, Hong Kong, China\\
    \textsuperscript{\rm 3}Institute of Information Engineering, Chinese Academy of Sciences, Beijing, China\\
    \textsuperscript{\rm 4}Universiti Sains Malaysia, Penang, Malaysia\\
    \textsuperscript{\rm 5}Xi'an Jiaotong University, Xi'an, China\\
    haojianhuang@connect.hku.hk, qinchuanyu24@mails.ucas.ac.cn,\\
    sun.010@163.com, hezhongj\_1@163.com
}

\usepackage{bibentry}

\begin{document}

\maketitle

\begin{abstract}
Multi-view classification (MVC) faces inherent challenges due to domain gaps and inconsistencies across different views, often resulting in uncertainties during the fusion process. While Evidential Deep Learning (EDL) has been effective in addressing view uncertainty, existing methods predominantly rely on the Dempster-Shafer combination rule, which is sensitive to conflicting evidence and often neglects the critical role of neighborhood structures within multi-view data. To address these limitations, we propose a \underline{T}rusted \underline{U}nified Feature-\underline{N}\underline{E}ighborhood \underline{D}ynamics (\textbf{TUNED}) model for robust MVC. This method effectively integrates local and global feature-neighborhood (F-N) structures for robust decision-making. Specifically, we begin by extracting local F-N structures within each view. To further mitigate potential uncertainties and conflicts in multi-view fusion, we employ a selective Markov random field that adaptively manages cross-view neighborhood dependencies. Additionally, we employ a shared parameterized evidence extractor that learns global consensus conditioned on local F-N structures, thereby enhancing the global integration of multi-view features. Experiments on benchmark datasets show that our method improves accuracy and robustness over existing approaches, particularly in scenarios with high uncertainty and conflicting views. The code will be made available at \textcolor{red}{https://github.com/JethroJames/TUNED}.
\end{abstract}

%
\section{Introduction}
\label{sec:intro}
Multi-view classification (MVC) has become a prominent research area due to the increasing availability of data from diverse sources. However, the existing MVC approaches predominantly encode multi-view features into a low-dimensional feature space through a feature bottleneck for fusion and classification. This straightforward fusion process often encounters inherent challenges, such as domain gaps and varying information content across views, which can introduce latent uncertainties during the fusion stage. To address this, many methods have explored measuring the quality of individual view representations to achieve discriminative multi-view fusion. Notably, Evidential Deep Learning (EDL), which stems from subjective logic~\cite{josang2016subjective} and Dempster-Shafer theory~\cite{dempster1968generalization}, has demonstrated effectiveness in assessing view representation uncertainty across various studies~\cite{huang2024crest,bao2021evidential,yu2023adaptive, qin2022deep, shao2024dual,xu2024reliable,yue2024evidential}.
\begin{figure}[H]
    \centering
    \includegraphics[width=\linewidth]{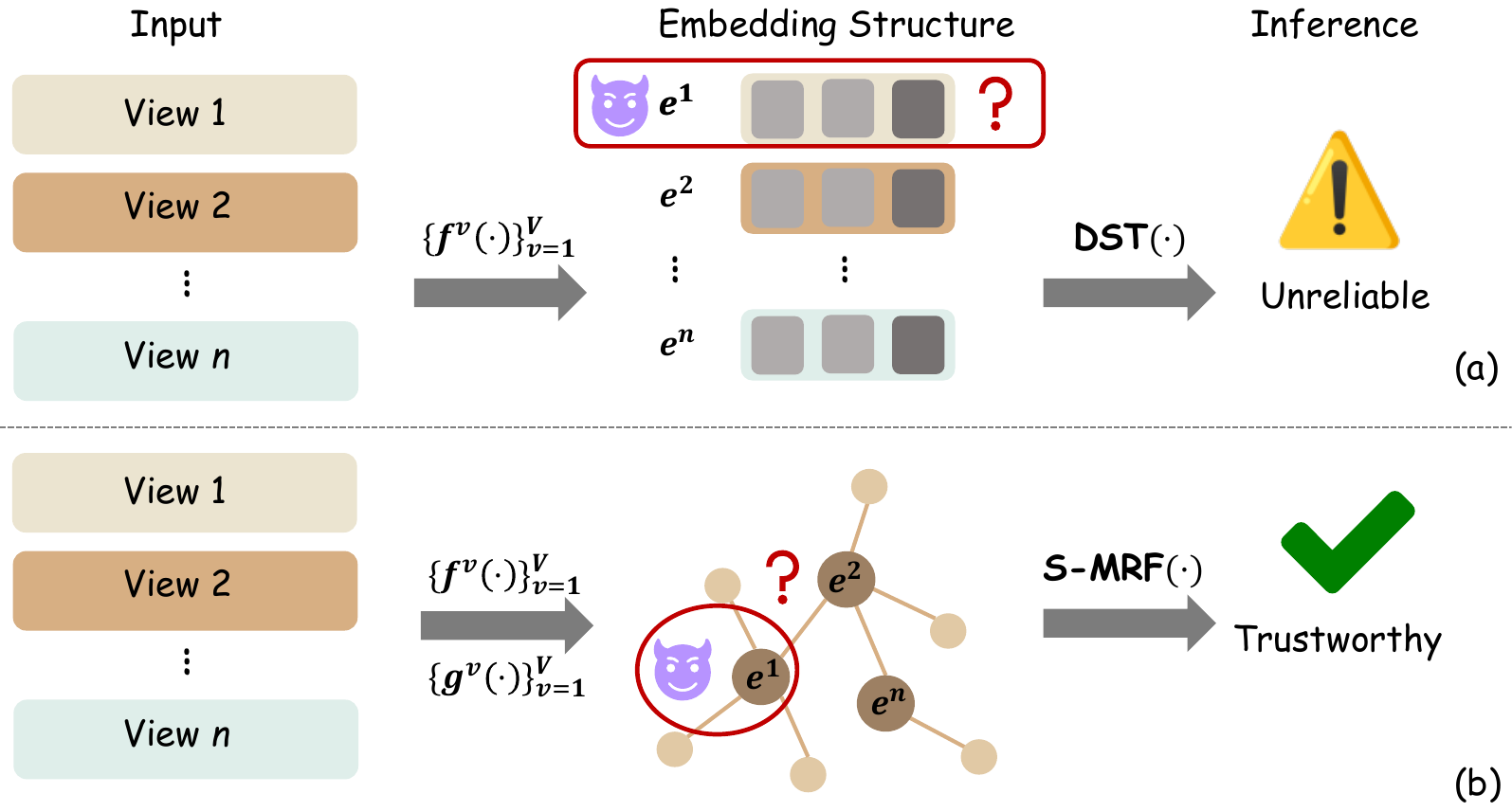}
    \vspace{-2mm}
    \caption{Comparison of EDL-based methods. (a) Conventional models use DNNs \( \{ f^v (\cdot) \}_{v=1}^V \) for evidence extraction, neglecting view-specific neighborhood structures and relying on the Dempster-Shafer theory (DST) for direct fusion, which can be unreliable with conflicting views. (b) Our method incorporates feature-neighborhood information via DNNs \( \{ f^v (\cdot) \}_{v=1}^V \) and GNNs \( \{ g^v (\cdot) \}_{v=1}^V \). The proposed Selective Markov Random Field (S-MRF) module dynamically fuses evidence, improving inference reliability without the need for hand-crafted loss functions.} 
    \label{fig:comp}
    \vspace{-3mm}
\end{figure}

Nevertheless, current EDL-based multi-view learning methods heavily rely on the Dempster-Shafer fusion framework. While theoretically robust, this framework is highly sensitive to conflicting evidence, where even a single contradictory source can produce anomalous results, ultimately compromising inference performance~\cite{huang2023belief,xiao2019multi}. Additionally, these methods often overlook the critical importance of neighborhood structure in multi-view features, leading to suboptimal performance in downstream tasks. For instance, in facial expression recognition, focusing solely on isolated view features can lead to misinterpretations due to variations such as lighting or angles. Without effectively leveraging both local and global neighborhood relationships, models struggle to capture the full context of the data, resulting in a less reliable fusion process. Unfortunately, existing EDL-based methods do not effectively leverage both local and global neighborhood relationships, which hinders their ability to fully capture the context of the data and leads to a less reliable fusion process.

To address these challenges, we propose a model for robust MVC method called \underline{T}rusted \underline{U}nified Feature-\underline{N}\underline{E}ighborhood \underline{D}ynamics (\textbf{TUNED}). To be specific, in the feature extraction stage, our method not only captures view-specific information but also integrates local neighborhood structures within each view. During the fusion stage, we simultaneously consider global neighborhood structures and the learning of cross-view dependencies, allowing for a more comprehensive understanding of the relationships between different views. To further refine this process, we introduce a dynamic Markov random field that adaptively accounts for cross-view relationships, suppressing views with lower coherence and learning optimal weights for each view. Moreover, We employ a parameterized evidence extractor to learn a collaborative evidence distribution. From this distribution, we sample evidence and perform a locally conditioned fusion with the evidence from each view. This approach enhances the joint learning of local and global feature-neighborhood structures, facilitating more effective and conflict-resilient integration across multiple views. Our contributions are threefold as follows:
\begin{itemize}
    \item We introduce a novel MVC framework that seamlessly integrates both local and global neighborhood structures. This approach enhances the robustness of feature extraction and fusion processes by effectively capturing cross-view dependencies and mitigating potential conflicts.
    \item We develop a selective Markov random field (S-MRF) model combined with a parameterized evidence extractor, which adaptively learns and fuses collaborative evidence across multiple views. This dual-layered evidence integration significantly improves the model’s ability to handle complex, heterogeneous data sources.
    \item Extensive experiments on benchmark datasets demonstrate that our method consistently outperforms state-of-the-art approaches, achieving superior accuracy and robustness in MVC tasks, particularly in scenarios with high levels of uncertainty and conflicts.
\end{itemize}
\begin{figure*}[h]
    \centering
    \includegraphics[width=\linewidth]{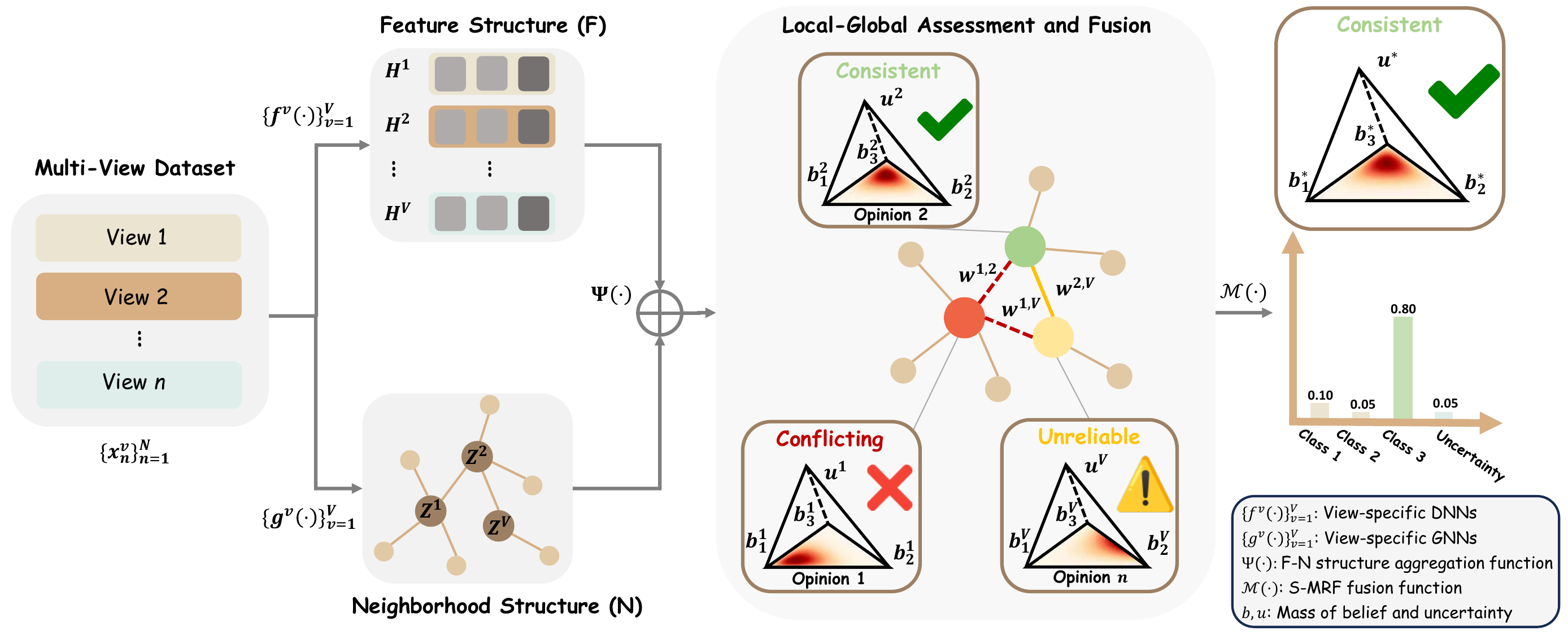}
    \vspace{-2mm}
    \caption{Illustration of the TUNED workflow, which comprises two stages. In the first stage, local view-specific feature-neighborhood (F-N) structures are extracted and fused to obtain evidence from the multi-view dataset. In the second stage, this evidence is integrated with global consensus F-N features through joint learning. The final step involves local-global evidence evaluation via EDL, followed by evidence fusion using the S-MRF module, leading to reliable inference outcomes.}
    \label{fig:workflow}
    \vspace{-3mm}
\end{figure*}
\section{Related work}
\subsection{Conflictive multi-view learning}
Most conflictive multi-view learning approaches focus on eliminating conflicting data instances. One line of work is based on multi-view outlier detection, which identifies outliers exhibiting abnormal behavior across multiple views. These methods are generally categorized into two types: cluster-based \cite{Huang_Ren_Pu_Huang_Xu_He_2023,zhang2023let,liu2024adaptive} and self-representation-based \cite{wang2019adversarial,wen2023highly}. Cluster-based methods perform clustering independently in each view and generate affinity vectors for each instance \cite{marcos2013clustering, zhao2017consensus}. Outliers are then detected by comparing these vectors across different views. In contrast, self-representation-based methods detect outliers by evaluating their difficulty in being represented by the normal views \cite{hou2020fast}.

Another approach focuses on partially view-aligned multi-view learning \cite{zhang2021late,wen2023unpaired}. Early work \cite{lampert2010weakly} introduced weakly-paired maximum covariance analysis to address challenges posed by unaligned data. More recently, \cite{huang2020partially} employed a differentiable variant of the Hungarian algorithm to align unaligned data. Subsequent work proposed noise-robust contrastive learning \cite{Yang_2021_CVPR,li2022positive,qin2024noisy} to compute alignment matrices.

However, these conflictive multi-view learning methods primarily aim to eliminate conflicting instances to achieve coordinated decisions across views. In scenarios like adversarial attack defense, making careful decisions in the presence of conflicting instances is crucial. To address this, recent methods have leveraged Evidential Deep Learning's (EDL) uncertainty measurement capabilities, quantifying both the conflict degree of among global proxy views~\cite{liu2023safe,Liu_Yue_Chen_Denoeux_2022, xu2024reliable} or local intra-view dissonance~\cite{yue2024evidential}. These methods seek to coordinate multiple views for decision-making without discarding conflicting instances. However, they universally rely on the Dempster-Shafer evidence fusion framework, which is highly sensitive to conflicting views~\cite{xiao2019multi,huang2023belief}. Without hand-crafted loss functions, this framework often yields counterintuitive and unreliable results, degrading task performance. To overcome these limitations, we propose a novel multi-view fusion framework that introduces a global-local F-N joint learning paradigm, further mitigating multi-view conflicts and enabling more reliable decision-making.

\subsection{Uncertainty-based deep learning}
Deep neural networks have made significant strides across various tasks, yet they often struggle to quantify uncertainty, especially when dealing with noisy or low-quality data~\cite{wen2023deep,chen2024finecliper}. To address these uncertainties, several approaches have been developed, including deterministic methods~\cite{sensoy2018evidential}, Bayesian neural networks~\cite{gal2016dropout}, ensemble methods~\cite{lakshminarayanan2017simple}, and test-time augmentation~\cite{lyzhov2020greedy}. These methods offer different mechanisms for uncertainty estimation, either by enhancing model robustness through probabilistic modeling or by using multiple predictions or augmentations for more accurate uncertainty quantification.

Recently, EDL, a prominent deterministic approach evolving from subjective logic~\cite{josang2016subjective} and Dempster-Shafer theory (DST)~\cite{dempster1968generalization}, has shown promise in uncertainty estimation. EDL has been successfully applied to tasks like MVC~\cite{han2021trusted,han2022trusted,Liu_Yue_Chen_Denoeux_2022,liu2023safe,xu2024reliable,huang2024evidential}, zero-shot learning~\cite{huang2024crest}, cross-modal retrieval~\cite{qin2022deep}, and action sequence localization~\cite{Chen2023TAL,Gao2023TAL,ma2024beyond}, demonstrating its versatility. In MVC, the pioneering work~\cite{han2021trusted} introduced the Dempster-Shafer combination rule, sparking a series of EDL-based studies. However, most of these methods assume that uncertainty will decrease after evidence fusion. The Dempster-Shafer rule, being highly sensitive to conflicting views, can produce counterintuitive and unreliable results when faced with conflicting evidence~\cite{xiao2019multi,huang2023belief}. Despite recent efforts to improve global coordination and local consistency through hand-crafted loss functions, the limitations of the Dempster-Shafer rule still hinder EDL's progress in MVC. To this end, we propose a novel fusion framework that mitigates the impact of conflicting views without relying on complex similarity measures or intricate loss designs.

\section{Methodology}
This section presents our proposed method for enhancing EDL through robust evidence fusion in conflictive MVC using the S-MRF. We begin by formally defining the problem, followed by a detailed explanation of how our model incorporates view-specific local F-N structures during the evidence extraction phase. Subsequently, we describe the use of S-MRF to suppress inter-view conflict dependencies, enabling the joint learning of local-global F-N structures. We then outline the loss functions involved in our approach. Finally, we discuss the factors contributing to the effectiveness of our method.

\subsection{Problem Definition}
In MVC, we are given a dataset \( \mathcal{D} = \{ \mathcal{X}_i, \boldsymbol{y}_i \}_{i=1}^n \), where \( n \) denotes the number of samples. Each sample \( \mathcal{X}_i \) consists of multiple views, represented as \( \mathcal{X}_i = \{ \boldsymbol{x}_i^v \}_{v=1}^V \), where \( V \) is the total number of views, and \( \boldsymbol{x}_i^v \in \mathbb{R}^{d_v} \) represents the feature vector of the \( i \)-th sample in the \( v \)-th view. The corresponding label for each sample is denoted by \( \boldsymbol{y}_i \in \mathcal{Y} \), where \( \mathcal{Y} \) is the set of possible labels. The target of MVC is to learn a model \( F_{\theta}: \mathcal{X} \rightarrow \mathcal{Y} \) that accurately predicts the label \( \boldsymbol{y}_i \) for an unseen sample \( \mathcal{X}_i \) by effectively integrating information from all available views \( \{ \boldsymbol{x}_i^v \}_{v=1}^V \). The challenge lies in leveraging the complementary information from each view while handling potential inconsistencies or conflicts among the views, ultimately improving the overall classification performance.

\subsection{Local F-N structure extraction}
\subsubsection{Local feature structure embedding.} We simply extract local features from each view using view-specific DNNs. Specifically, for each view \( v \), we extract feature representations \(\boldsymbol{\mathit{h}}_{n}^{v}\) from the input data using DNNs \( \{ f^v (\cdot) \}_{v=1}^V \). Given an input feature vector \( \boldsymbol{\mathit{x}}_n^v \), the corresponding DNN \( f^v(\cdot) \) maps this to a feature vector:
\begin{equation}
    \boldsymbol{\mathit{h}}_{n}^{v} = f^v(\boldsymbol{\mathit{x}}_n^v; \theta^v)
\end{equation}
where \( \theta^v \) denotes the parameters of the DNN for view \( v \). These local feature representations inherently capture the local neighborhood structure within each view, providing critical information for subsequent integration. 
\subsubsection{Local neighborhood structure embedding.} In multi-view data analysis, traditional feature extraction methods often fail to account for the inherent local neighborhood structures among samples. This omission is particularly problematic in datasets with non-Euclidean characteristics, such as social or biological networks, where intricate patterns are prevalent. Capturing both local and global relationships is crucial for fully understanding the data's complexity. To effectively capture the complex local neighborhood relationships within multi-view data, we employ the Clustering-with-Adaptive-Neighbors (CAN)~\cite{nie2014clustering} approach to construct an adaptive adjacency matrix for our Graph Convolutional Network (GCN). This method dynamically determines the neighborhood structure for each sample, enabling the model to robustly reflect both local and global data characteristics.

Given a feature matrix \( X^v \in \mathbb{R}^{d_v \times n} \), where \( d_v \) represents the feature dimension for view \( v \) and \( n \) represents the number of samples, we first compute the pairwise distance matrix \( D \in \mathbb{R}^{n \times n} \), where each element \( D_{ij} = \|\boldsymbol{x}_i^v - \boldsymbol{x}_j^v\|_2^2 \) represents the squared distance between samples \( \boldsymbol{x}_i^v \) and \( \boldsymbol{x}_j^v \). Next, we sort the distances for each sample to identify its top \( k \) nearest neighbors. Let \( \varphi (\boldsymbol{x}_i^v) \) denote the distance to the \( k \)-th nearest neighbor for sample \( \boldsymbol{x}_i^v \). We then compute a weight matrix \( \boldsymbol{\mathcal{M}} \) using the formula \( T_{ij} = \varphi (\boldsymbol{x}_i^v) - D_{ij} \), which balances the influence of each neighbor as follows:
\begin{equation}
\mathcal{M}_{ij} = \max\left(0, \frac{T_{ij}}{k \cdot \varphi(\boldsymbol{x}_i^v) - \sum_{j \in N_i} D_{ij}}\right),
\end{equation}
where \( N_i \) represents the set of \( {k} \) nearest neighbors for sample \( \boldsymbol{x}_i^v \). This adaptive weighting strategy ensures that each sample’s neighborhood captures the most relevant local relationships, leading to a more representative and robust graph structure. The symmetrized adjacency matrix \( \mathbf{A} \) is then computed as:
\begin{equation}
\mathbf{A}^{v} = \frac{\boldsymbol{\mathcal{M}}^{v} + (\boldsymbol{\mathcal{M}}^{v})^\top}{2},
\end{equation}
where $\mathbf{A}$ is used as the input for the view-specific GCN, and the feature extraction process is defined as below:
\begin{equation}
\mathbf{Q}^{v}_{l+1} = \sigma(\mathbf{\hat{A}}^{v} \mathbf{Q}^{v}_l \mathbf{W}^{v}_l),
\end{equation}
where \(\mathbf{\hat{A}}^{v}\) is the normalized adjacency matrix, \(\sigma\) is the activation function (e.g., ReLU, Softplus), and \(\mathbf{W}^{v}_l\) are the learnable weights at layer \(l\).

\subsection{Neighborhood-aware evidential deep learning}
In multi-view learning, capturing the complex relationships among data points is essential for accurate classification. Traditional methods often overlook neighborhood structures, leading to suboptimal integration of multi-view data, especially in non-Euclidean spaces like social or biological networks. To address this, we propose neighborhood-aware EDL, which explicitly incorporates neighborhood structure. By fusing local-global F-N structures, our approach enhances the model's ability to account for dependencies among data points and achieve more reliable and robust classification.

\subsubsection{Local evidence extraction.}
In our framework, the neighborhood-aware features are fused using a function \(\Psi(\cdot)\) (See Supplementary Material) with an activation layer (\emph{e.g.}, ReLU, Softplus) to obtain non-negative output as evidence for each view, so that we can obtain the parameters of the Dirichlet distribution. For a classification problem with \( K \) classes, the evidence \(\boldsymbol{\mathit{e}}_{n}^{v}\) for an instance \( \mathbf{x}_{n}^{v} \) in view \( v \) is represented by an opinion \(\boldsymbol{\omega}^{v} = (\boldsymbol{\mathit{b}}^{v}, \mathit{u}^{v})\), where \( \boldsymbol{\mathit{b}}^{v} = (\mathit{b}_{1}^{v},\dots,\mathit{b}_{K}^{v})^{\top} \) is the belief mass vector, which assigns belief values to each class based on the evidence and \( \mathit{u} \) is the uncertainty mass, representing the degree of uncertainty in the evidence. These components are constrained by the following relationship:
\begin{equation}
\sum_{k=1}^{K} \mathit{b}_{k}^{v} + \mathit{u}^{v} = 1, \quad \text{where} \quad \mathit{b}_{k}^{v} \geq 0 \quad \text{and} \quad \mathit{u}^{v} \geq 0.
\end{equation}

The belief \( \boldsymbol{\mathit{b}}^{v} \) and uncertainty \( \mathit{u}^{v} \) are linked to the Dirichlet distribution, which models second-order uncertainty. The Dirichlet probability density function (PDF) is defined as:
\begin{equation}
\mathit{D}(\mathbf{p}|\boldsymbol{\alpha}^{v} ) =  \left\{
\begin{matrix}
\frac{1}{\mathit{B}(\boldsymbol{\alpha})}  \prod_{k=1}^{K}\mathit{p}_{k}^{\mathit{\alpha}^{v}_{k}-1}, & \text{for } \mathbf{p}\in \mathcal{S}_{K}^{v}, \\
0, & \text{otherwise},
\end{matrix}
\right.
\end{equation}

where \(\boldsymbol{\mathbf{p}}^{v} = (\mathit{p}_{1}^{v},\dots,\mathit{p}^{v}_{K})^{\top}\) is the probability assigned to each class \( k \), and \(\boldsymbol{\mathit{\alpha}} = (\mathit{\alpha}_{1},\dots,\mathit{\alpha}_{K})^{\top }\) are the Dirichlet parameters. The simplex \(\mathcal{S}_{K}\) is defined as:

\begin{equation}
\mathcal{S}_{K}^{v} = \left \{ \mathbf{p}^{v} \middle| \sum_{k=1}^{K}\mathit{p}_{k}^{v}=1, \quad 0 \le \mathit{p}_{k}^{v} \le 1 \text{ for all } k \right \}.
\end{equation}

The view-specific Dirichlet parameters \(\boldsymbol{\alpha}^{v}\) are computed as \(\boldsymbol{\alpha}^{v} = \boldsymbol{\mathit{e}}^{v} + \mathbf{1}\), ensuring non-sparsity. The belief and uncertainty are related to these parameters by:

\begin{equation}
\mathit{b_{k}^{v}} = \frac{\mathit{e}_{k}^{v}}{S^{v}} = \frac{\mathit{\alpha_{k}^{v}}-1}{S^{v}}, \quad \mathit{u}^{v} = \frac{K}{S^{v}},
\end{equation}
where \( S^{v} = \sum_{k=1}^{K} (\mathit{e}_{k}^{v}+1) = \sum_{k=1}^{K} \mathit{\alpha}_{k}^{v} \). The uncertainty \( \mathit{u} \) is inversely proportional to the total evidence, indicating that as the evidence increases, the uncertainty decreases.

After obtaining the view-specific evidence, this step reflects the confidence of the local proxy in classifying based on the F-N structures. This evidence is then further integrated with global F-N structures in a joint learning manner, setting the stage for robust multi-view evidences fusion.

\subsubsection{Global consensus evidence extraction.}
To resolve conflicts between views and establish a robust multi-view consensus, we introduce a consensus evidence extraction mechanism. This mechanism samples a consensus evidence vector \(\mathbf{e}_n^{\text{cons}}\) from a shared Dirichlet distribution and conditions it on the local F-N structures through a fusion function \(\Phi(\cdot)\), which integrates the sampled consensus evidence with the view-specific evidence vectors as follows:
\begin{equation}
\label{eq:view_evidence}
\tilde{\mathbf{e}}_{n}^v = \mathbf{e}_n^v + \Phi\left(\mathbf{e}_n^{\diamond}, \mathbf{e}_n^v \right),
\end{equation}
where \(\mathbf{e}_n^{\diamond } \sim \text{Dir}(\boldsymbol{\alpha}_n^{\diamond })\) is the evidence vector randomly sampled from the shared Dirichlet distribution parameterized by \(\boldsymbol{\alpha}_n^{\diamond}\). The fusion function \(\Phi(\cdot)\) (\emph{e.g.} Self-Attention, cosine similarity \emph{etc.}) conditions the consensus evidence on the local F-N structures, facilitating robust global learning of multi-view consensus evidence. This approach ensures that the consensus evidence effectively reflects patterns across all views, while also being adaptable to local variations.

\subsection{Conflictive evidence aggregation}
Conflicts among views are prevalent yet often overlooked by existing methods. Directly integrating all views can lead to performance degradation, as seen in frameworks like the Dempster-Shafer Theory, where minor local conflicts can cause global fusion failures. To address this, we propose the S-MRF model, which adaptively selects highly relevant views for evidence fusion. This model mitigates the impact of local conflicts on global neighborhood-aware evidence integration, ensuring robust and reliable inference.

\subsubsection{Selective Markov random field.}
Specifically, the S-MRF model represents the interaction among views through an energy function, which considers both unary and pairwise potentials. The energy function is defined as below:
\begin{equation}
\mathrm{E}(\mathbf{Z}) = \sum_{i} \phi(\mathbf{z}_i) + \sum_{i,j} \sigma (\mathbf{z}_i, \mathbf{z}_j),
\end{equation}
where \(\phi(\mathbf{z}_i)\) represents the unary potential representation of view \(i\), and \(\sigma (\mathbf{z}_i, \mathbf{z}_j)\) represents the pairwise potential representation between views \(i\) and \(j\). In the S-MRF model, each view \(v\) is associated with an evidence vector \(\mathbf{e}_v\). A graph \(\mathcal{G} = (\mathcal{V}, \mathcal{E})\) is constructed, where \(\mathcal{V}\) corresponds to the set of views, and \(\mathcal{E}\) represents the set of edges determined by the similarity between views. The edge weight \(w_{i,j}\) reflects the similarity between views \(i\) and \(j\), calculated using metrics such as cosine similarity or other suitable measures.

Edges are included in the graph if they meet a threshold \(\tau\), ensuring that only significant and relevant connections are retained, thus addressing inter-view conflicts:
\begin{equation}
w_{i,j} \geq \tau \cdot w_{\text{max}},
\end{equation}
where \(w_{\text{max}}\) is the maximum edge weight in the initial connectivity matrix. 

\subsubsection{Local-global evidence aggregation.}
According to equation~(\ref{eq:view_evidence}), we denote view $v$ evidence as $\mathbf{\tilde{E}}_v$, which accounts for local-global F-N structures.  The aggregated evidence \(\mathbf{\tilde{E}}_{\text{agg}}\) from these selectively connected views is computed as:
\begin{equation}
\mathbf{\tilde{E}}_{\text{agg}} = \sum_{(i,j) \in \mathcal{E}} w_{i,j} \cdot \mathbf{\tilde{E}}_j.
\end{equation}

By normalizing the contributions, the S-MRF model effectively combines evidence from views that are contextually significant, reducing the impact of irrelevant or conflicting views on the overall integration process. This selective approach enhances integration efficiency and improves the interpretability of the resulting evidence fusion.



\subsection{Loss function}
\subsubsection{Neighborhood-aware EDL loss.}
For each instance \(\{ \mathbf{x}_{n}^v \}_{v=1}^V\), we fuse the extracted feature structure \( \mathbf{h}_{n}^{v} \) and neighborhood structure \( \mathbf{q}_{n}^{v} \) using a F-N aggregation function \( \Psi(\cdot) \) to obtain evidence \( \mathbf{e}_{n}^{v} \) for each view:
\begin{equation}
\mathbf{e}_{n}^{v} = \Psi(\mathbf{h}_{n}^{v}, \mathbf{q}_{n}^{v}).
\end{equation}

This evidence vector \( \mathbf{e}_{n}^{v} \) is then used to parameterize the Dirichlet distribution $\boldsymbol{\alpha}_{n}^{v} = \mathbf{e}_{n}^{v} + \mathbf{1}$. Then we adapt the conventional cross-entropy loss to integrate the evidence and uncertainty from each view, enhancing the robustness of the classification process as follows.
\begin{align}
\nonumber \mathit{\mathcal{L}_{ace}}(\boldsymbol{\alpha}_{n}^{v}) &= \int \left [ \sum_{j=1}^{K}-y_{nj}\log_{}{p_{nj}}\right ] \frac{\textstyle \prod_{j=1}^{K}p_{nj}^{{\alpha}_{nj}-1}}{B\left ( \boldsymbol{\alpha}_{n}^{v} \right)}d\mathbf{p}_{n}\\ 
&=\sum_{j=1}^{K}y_{nj}(\psi(S_{n}^{v})-\psi({\alpha}_{nj}^{v})),
\end{align}
where $\psi(\cdot)$ is the digamma function. The above loss function does not guarantee that the evidence generated by the incorrect labels is lower. To address this issue, we can introduce an additional term in the loss function, namely the Kullback-Leibler (KL) divergence:
\begin{align}
\mathcal{L}_{KL}(\boldsymbol{\alpha}_{n}) 
&= KL \left[D(\boldsymbol{p}_{n}|\tilde{\boldsymbol{\alpha}}_{n})\parallel D(\boldsymbol{p}_{n}|\mathbf{1})\right] \\ \nonumber
&= \log_{}{(\frac{\Gamma( {\textstyle \sum_{k=1}^{K}}\tilde{\alpha}_{nk})}{\Gamma(K) {\textstyle \prod_{k=1}^{K}\Gamma(\tilde{\alpha}_{nk})} })} \\ \nonumber
&+ \sum_{k=1}^{K}(\tilde{\alpha}_{nk}-1)\left [ \psi(\tilde{\alpha}_{nk})-\psi (\sum_{j=1}^{K}\tilde{\alpha}_{nj}) \right ],
\end{align}
where $D(\mathbf{p}_{n}|\mathbf{1})$ is the uniform Dirichlet distribution, $\tilde{\boldsymbol{\alpha}}_{n}^v = \mathbf{y}_{n} + (\mathbf{1}-\mathbf{y}_{n}) \odot \boldsymbol{\alpha}_{n}^v$ is the Dirichlet parameters after removal of the non-misleading evidence from predicted parameters for the $n$-th instance, and $\Gamma(\cdot)$ is the gamma function. 

Therefore, given the Dirichlet distribution with parameter $\alpha_{n}^{v}$ for the $n$-th instance, the loss is:
\begin{align}
\mathcal{L}_{acc}(\boldsymbol{\alpha}_{n}^v) = \mathcal{L}_{ace}(\boldsymbol{\alpha}_{n}^v) \ + \ \lambda_{s}\mathcal{L}_{KL}(\boldsymbol{\alpha}_{n}^v),
\end{align}
where \(\lambda_{s} = \min(1.0, s/\mathcal{T}) \in \left[ 0, 1 \right]\) is the annealing coefficient, with \(t\) representing the current training epoch and \(\mathcal{T}\) the total number of annealing steps. This gradual increase in the influence of KL divergence in the loss function helps prevent premature convergence of misclassified instances towards a uniform distribution.

\subsubsection{Conflict-aware aggregation loss.}
In the proposed S-MRF module, each graph node represents evidence derived from different views. To achieve conflict-aware MVC, it is essential to harmonize the evidence across these views. This is accomplished by a combined loss function that simultaneously maximizes the pairwise similarity between evidence and minimize the deviation of each evidence from their mean. Moreover, we introduce a regularizer to prevent node degradation and maintain diversity in the graph.

Specifically, to ensure global consistency and inter-view collaboration, we introduce a unified consistency loss \(\mathcal{L}_{\text{con}}\), which combines both the similarity encouragement in terms of statistical property and regularization components:
\begin{equation}
\begin{split}
\mathcal{L}_{\text{con}} = &- \frac{2}{V(V-1)}  \sum_{i=1}^{V-1} \sum_{j=i+1}^{V} \frac{\mathbf{e}_i \cdot \mathbf{e}_j}{\|\mathbf{e}_i\| \|\mathbf{e}_j\|} \\
&+  \frac{1}{VN} \sum_{v=1}^{V} \sum_{n=1}^{N} \|\mathbf{e}_n^v - \boldsymbol{\mu}^v\|_2^2 -  \sum_{v=1}^{V} \|\mathbf{E}^v\|_\text{F}^2,   
\end{split}
\end{equation}
where $\boldsymbol{\mu}^v$ represents the mean evidence across all samples within view $v$, while $\mathbf{\mathbf{E}^v}$ denotes the $v$-th view evidence.

To sum up, the overall loss function for a specific instance $\{ \mathbf{x}_{n}^v \}_{v=1}^V$ can be calculated as:
\begin{equation}
\vspace{-2mm}
\mathcal{L} =  \frac{1}{V}\sum_{v=1}^{V}\mathcal{L}_{acc}(\boldsymbol{\alpha}_{n}^v) + \lambda_t\mathcal{L}_{\text{con}}
\end{equation}
The model optimization process and extensive discussion can be found in \textbf{Supplementary Material}.
\vspace{-2mm}
\section{Experiment}
\subsection{Experimental setup}
\subsubsection{Datasets.}
The overview of the datasets used in our study is illustrated in Table~\ref{tab:datasets}, with more detailed descriptions provided in the \textbf{Supplementary Material}. In summary, we benchmarked our approach against state-of-the-art (SOTA) methods on eight multi-view datasets. The conflictive datasets were constructed by us based on these datasets, following the methodology outlined in~\cite{xu2024reliable}. Each method was run 10 times, and we report the average performance along with the standard deviation.
\begin{table}[htbp]
\centering
\caption{Overview of datasets used in experiments}
\vspace{-2mm}
\label{tab:datasets}
\small 
\begin{tabular}{@{}lccc@{}}
\toprule
\textbf{Dataset} & \textbf{Size} & \textbf{Classes} & \textbf{Dimensionality} \\ \midrule
PIE  & 680 & 68 & 484\textbar256\textbar279 \\
HandWritten & 2000 & 10 & 240\textbar76\textbar216\textbar47\textbar64\textbar6 \\
Scene15 & 4485 & 15 & 20\textbar59\textbar40 \\
Caltech101 & 8677 & 101 & 4096\textbar4096 \\
Animal& 10158& 50&4096\textbar4096 \\
ALOI & 10800 & 100 & 77\textbar13\textbar64\textbar125 \\
CUB & 11788 & 10 & 1024\textbar300 \\
NUS-WIDE-OBJECT & 30000 & 31 & 129\textbar74\textbar145\textbar226\textbar65 \\ \bottomrule
\end{tabular}
\vspace{-4mm}
\end{table}

\subsubsection{Compared methods.}
In this study, we compared our approach against eight SOTA methods. Among these, TMC~\cite{han2021trusted}, TMDL-OA~\cite{Liu_Yue_Chen_Denoeux_2022}, ECML~\cite{xu2024reliable}, and RMVC~\cite{yue2024evidential} are EDL-based multi-view methods, while EDL~\cite{sensoy2018evidential} itself is a single-view method. Detailed descriptions of all the comparison methods can be found in the \textbf{Supplementary Material}.
\subsection{Quantative results}
\subsubsection{Comparison with SOTA.} Tables~\ref{tab:normal_acc} and~\ref{tab:conflict_acc} show the classification performance on normal and conflictive test sets. We can observe that: \textbf{(1)} On normal test sets, TUNED outperforms all other baselines on all datasets except for the HandWritten and Scene15 datasets. \textbf{(2)} When evaluating on conflictive test sets, the accuracy of all compared methods significantly decreases. However, TUNED achieves impressive results across all datasets, with particularly large improvements on the Scene15, Animal, and ALOI datasets (18\%, 37.73\%, and 15.4\% respectively). Additionally, we notice that on the Animal dataset, the conflictive views even achieve better results than on the normal test sets. This might be due to the sensitivity of the parameters being further amplified by the reduced number of views, a phenomenon that has been validated in subsequent studies.

\subsubsection{Ablation study.} Table~\ref{tab:ablation} presents the ablation study results of TUNED on the HandWritten and Scene datasets. The experimental results indicate that each component of our framework contributes positively to multi-view fusion. Moreover, compared to other fusion methods such as Arithmetic mean and DST, our MRF fusion demonstrates superior performance. We provide additional comparisons based on a broader set of metrics in the \textbf{Supplementary Material}.
\subsection{Qualitative results}

\begin{table*}[h]
\centering
\vspace{-2mm}
\caption{Accuracy (\%) on normal test sets. The best and the second best results are highlighted by \textbf{boldface} and \underline{underlined} respectively. $\bigtriangleup \%$ denotes the performance improvement of ours over the best baseline. NUS refers to ``NUS-WIDE-OBJECT''.}
\label{tab:normal_acc} 
\small
\begin{tabular*}{1.00\textwidth}{@{\extracolsep{\fill}}ccccccccccc}
\toprule
Method & PIE & HandWritten & Scene15 & Caltech101 & CUB & Animal & ALOI & NUS \\
\midrule
EDL & 86.25±0.89 & 96.90±0.16 & 52.76±0.54 & 73.35±1.73 & 86.22±0.36 & 84.30±1.76 & 37.87±5.13 & 22.33±0.64 \\
DCCAE & 81.96±1.04 & 95.45±0.35 & 74.62±1.52 & 89.56±0.41 & 85.39±1.36 & 82.72±1.38 & 87.20±0.14 & \underline{35.75±0.48} \\
CPM-Nets & 88.53±1.23 & 94.55±1.36 & 67.29±1.01 & 90.35±2.12 & 89.32±0.38 & 87.40±1.12 & 49.13±6.19 & 35.37±1.05 \\
DUA-Nets & 90.56±0.47 & 98.10±0.32 & 68.23±0.11 & 93.43±0.34 & 80.13±1.67 & 78.65±0.55 & 83.09±2.15 & 33.98±0.34 \\
TMC & 91.85±0.23 & 98.51±0.13 & 67.71±0.30 & 92.80±0.50 & 90.57±2.96 & 87.20±0.33 & 79.31±0.31 & 35.18±1.55 \\
TMDL-OA & 92.33±0.36 & \underline{99.25±0.45} & 75.57±0.02 & 94.63±0.04 & 95.43±0.20 & 87.05±0.28 & 65.20±0.18 & 34.39±0.44 \\
ECML & \underline{94.71±0.02} & \textbf{99.40±0.00} & \textbf{76.19±0.12} & \underline{95.36±0.38} & \underline{98.50±2.75} & 84.01±0.63 & 69.29± 0.11 & 34.04 ± 0.27 \\
RMVC & 91.18±0.24 & 98.51±0.04 & 73.05±0.24 & 88.73±0.60 & 93.18±0.47 & \underline{87.67±0.17} & \underline{87.38±0.67} & 34.68±0.32 \\
Ours & \textbf{96.83±0.01} & 99.20±0.23 & \underline{75.89±0.70} & \textbf{97.38±0.45} & \textbf{99.00±0.33} & \textbf{89.52 ± 0.13} & \textbf{88.93±0.35} & \textbf{37.46±0.25} \\
 $\bigtriangleup \%$  & 2.24 & -0.2 & -0.39 & 2.12 & 0.51 & 2.11 & 1.77 & 4.78 \\
\bottomrule
\end{tabular*}
\vspace{-1mm}
\end{table*}
           
\begin{table*}[h]
\centering
\vspace{-2mm}
\caption{Accuracy (\%) on conflictive test sets. The tags setting is the same as that in Table~\ref{tab:normal_acc}.}
\label{tab:conflict_acc}
\small
\begin{tabular*}{1.00\textwidth}
{@{\extracolsep{\fill}}ccccccccccc}
\toprule
Method & PIE & HandWritten & Scene15 & Caltech-101 & CUB & Animal & ALOI & NUS \\
\midrule
EDL & 21.76±0.67 & 57.25±0.49 & 14.28±0.24 & 55.74±0.12 & 53.75±0.42 & 30.71±0.27 & 25.05±3.10 & 18.07±0.28 \\
DCCAE & 26.89±1.10 & 82.85±0.38 & 25.97±2.86 & 60.90±2.32 & 63.57±1.28 & 64.30±2.11 & 75.12±0.43 & 32.12±0.52 \\
CPM-Nets & 53.19±1.17 & 83.34±1.07 & 29.63±1.12 & 66.54±2.89 & 68.82±0.17 & 64.83±0.35 & 36.29±5.02 & 29.20±0.81 \\
DUA-Nets & 56.45±1.75 & 87.16±0.34 & 26.18±1.31 & 75.19±2.34 & 60.53±1.17 & 62.46±1.12 & 69.07±2.50 & 31.82±0.43 \\
TMC & 61.65±1.03 & 92.76±0.15 & 42.27±1.61 & 90.16±2.50 & 73.37±2.16 & 64.85±1.19 & \underline{76.68±0.32} & \underline{33.76±2.16} \\
TMDL-OA & 68.16±0.34 & 93.05±0.05 & 48.42±1.02 & 90.63±2.05 & 74.43±0.26 & 64.62±0.15 & 62.90±0.12 & 32.44±0.26 \\
ECML & \underline{84.00±0.14} & 94.40±0.05 & \underline{56.97±0.52} & \underline{92.36±1.48} & \underline{76.50±1.15} & 62.67±0.81 & 64.91 ± 0.20 & 31.19 ± 0.22 \\
RMVC & 76.47±3.43 & \underline{94.75±0.75} & 49.83±2.23 & 80.56±0.71 & 72.78±0.42 & \underline{66.00±0.59} & 52.67±1.97 & 24.62±3.19 \\
Ours & \textbf{86.02±0.19} & \textbf{96.75± 0.55} & \textbf{67.22± 0.58} & \textbf{93.22±0.41} & \textbf{76.67±0.38} & \textbf{90.90± 0.18} & \textbf{88.49±0.29} & \textbf{34.09±0.14} \\
 $\bigtriangleup \%$ & 2.40 & 2.11 & 18.00 & 0.93 & 2.22 & 37.73 & 15.40 & 0.98 \\
\bottomrule
\end{tabular*}
\vspace{-2mm}
\end{table*}

\begin{table}[h]
\centering
\caption{Ablation study: Accuracy (\%) on HandWritten and Scene15 datasets under normal and conflict conditions.}
\vspace{-2mm}
\label{tab:ablation}
\small 
\begin{tabular}{lcccc} 
\toprule
Method & \multicolumn{2}{c}{HandWritten} & \multicolumn{2}{c}{Scene15} \\
\cmidrule(lr){2-3} \cmidrule(lr){4-5}
 & Normal & Conflict & Normal & Conflict \\
\midrule
w/o GCN & 95.75 & 95.75 & 69.23 & 64.21 \\
w/o consensus\_evidence & 96.50 & 94.25 & 70.79 & 64.88 \\
w/o $\mathcal{L}_{\text{con}}$ & 96.50 & 94.50 & 70.79 & 65.44 \\
w/ Average & 96.25 & 96.00 & 70.90 & 65.33 \\
w/ DST & 95.75 & 91.75 & 70.56 & 66.67 \\
w/ S-MRF & \textbf{99.20} & \textbf{96.75} & \textbf{75.89} & \textbf{67.22} \\
\bottomrule
\end{tabular}
\end{table}
\begin{figure}[h]
    \centering
    \includegraphics[width=\linewidth]{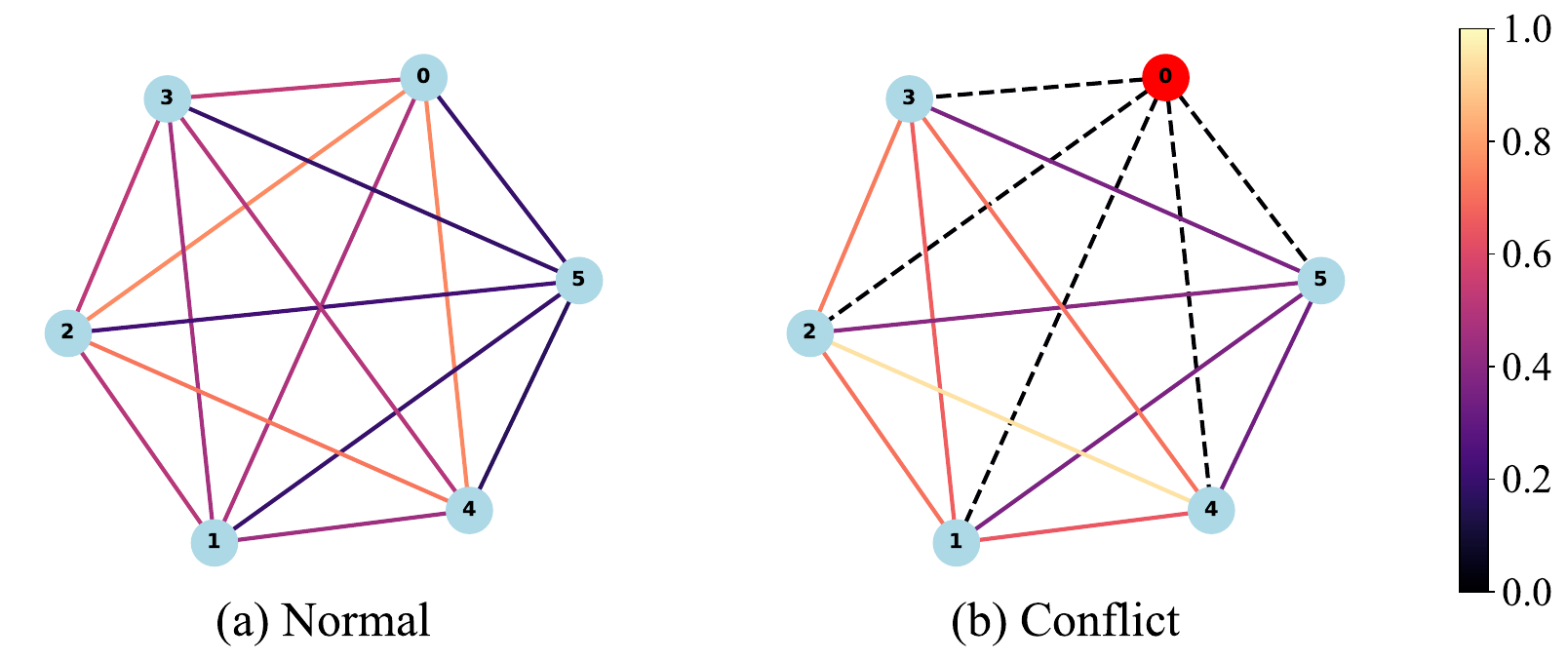}
    \vspace{-2mm}
    \caption{S-MRF weights for the HandWritten dataset. (a) Represents the weights under the normal setting, while (b) illustrates the weights under the conflict setting. And the red node indicates the view where conflict have been introduced.}
    \label{fig:MRF}
    \vspace{-4mm}
\end{figure}
\begin{figure}[H]
    \centering
    \includegraphics[width=\linewidth]{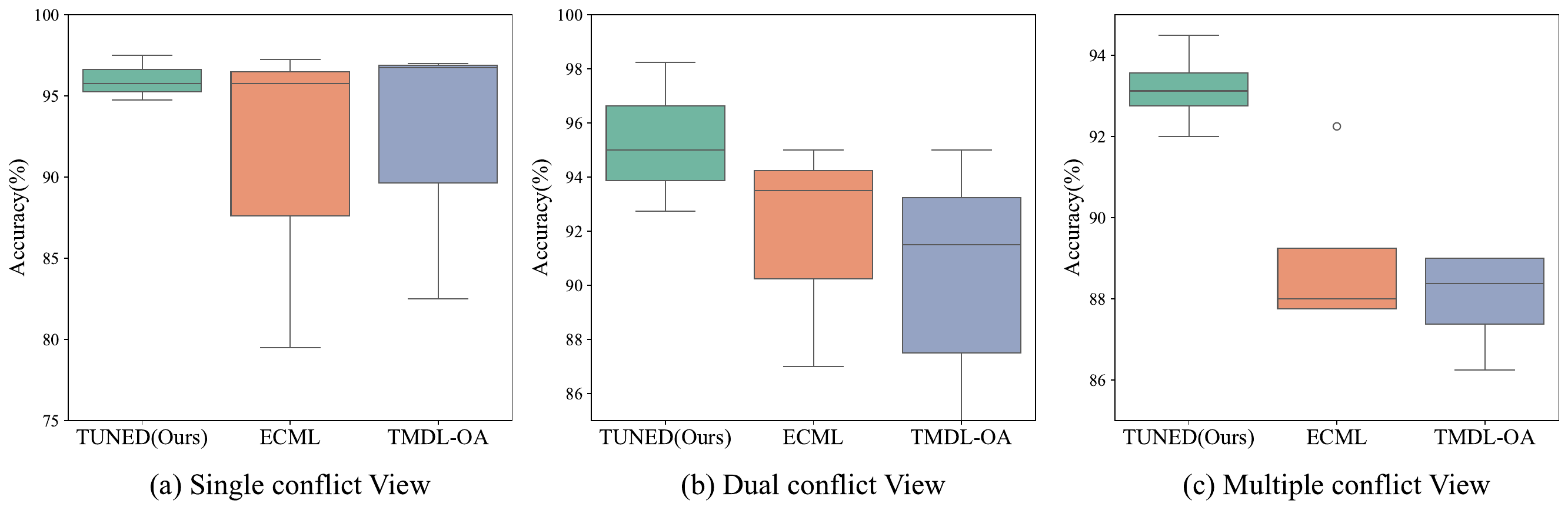}
    \vspace{-2mm}
    \caption{Robustness comparison: Impact of conflict views on model performance across different configurations.} 
    \label{fig:box}
    \vspace{-4mm}
\end{figure}
\begin{figure}[htbp]
    \centering
    \includegraphics[width=\linewidth]{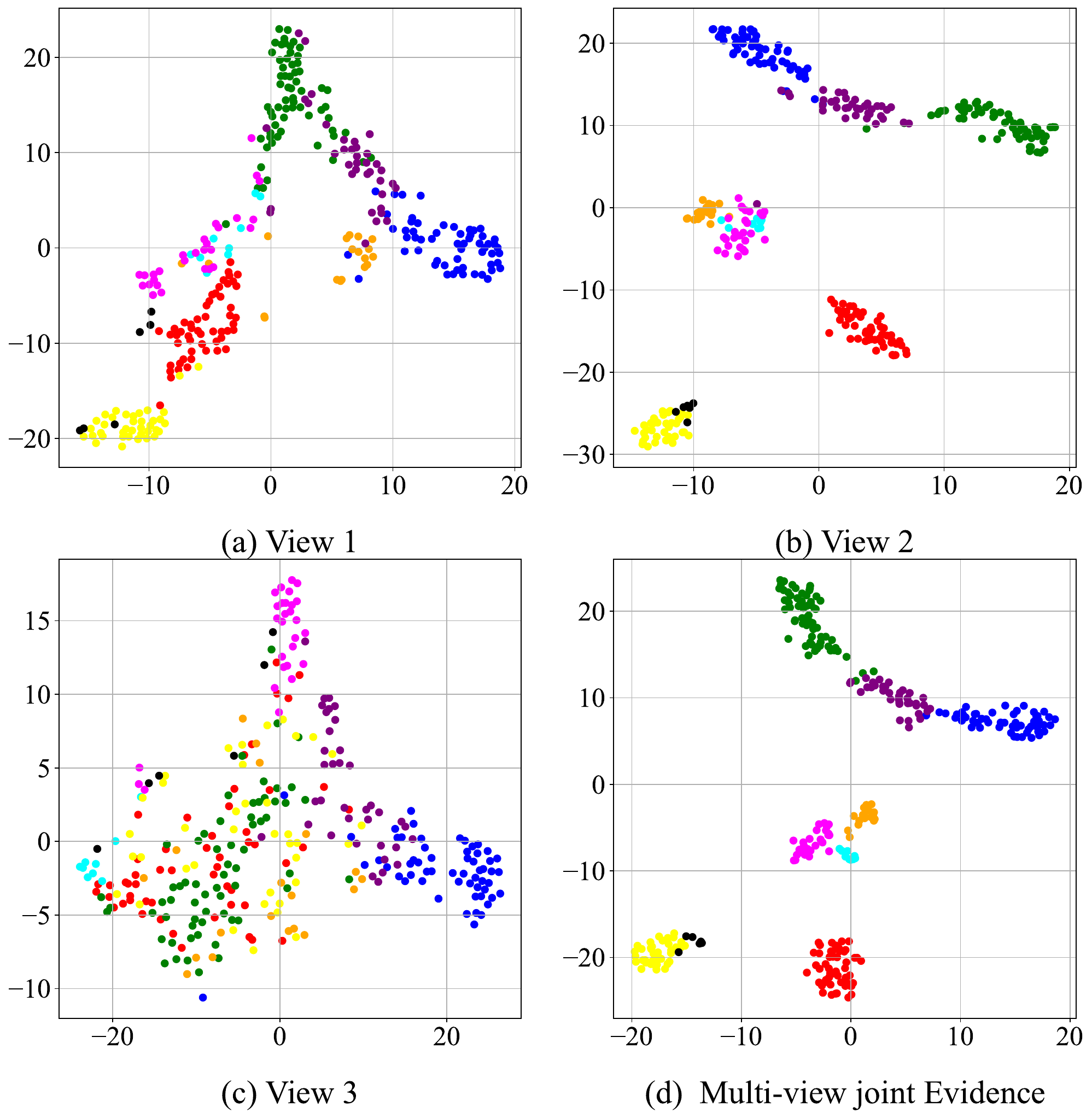}
    \vspace{-2mm}
    \caption{T-SNE visualization of Evidence. (a)-(c) show the t-SNE results for Evidence from three individual views. (d) illustrates the fused Evidence after t-SNE, highlighting the more cohesive data representation achieved through fusion.} 
    \label{fig:tsne}
    \vspace{-4mm}
\end{figure}
\subsubsection{Visualization of S-MRF.} The Figure~\ref{fig:MRF} illustrates the weight relationships between different views in the HandWritten dataset. The left and right panels show the conflict levels for normal and conflict instances, respectively. To create conflict, we modified the content of view 0, resulting in misalignment with the other views, and set the corresponding \(\tau\) to 0.7. The results demonstrate that, after training, the weights between the conflict view and the other views are reduced to zero. This clearly indicates that our fusion method effectively captures and mitigates the impact of conflict views on model integration.
\subsubsection{Robustness to conflict.} We evaluated TMDL-OA, ECML, and our TUNED model on the HandWritten dataset with varying numbers of noisy views in Figure~\ref{fig:box}. Conflict test sets were created by adding noise to single, paired, and multiple views among the six available. Results show that as more views are corrupted, accuracy generally decreases. However, our TUNED model consistently outperforms others, demonstrating lower variance and greater robustness, especially in scenarios with noise in single views, where ECML and TMDL-OA show significant instability. 
\subsubsection{T-SNE visualization.} We visualized the Evidence from three views and their fused representation on the Scene dataset after model training in Figure~\ref{fig:tsne}. The results demonstrate that MRF effectively captures inter-view relationships, producing a fused Evidence representation with more distinct class separations, thereby enhancing the discriminative power between categories.
\subsubsection{Parameters analysis.} We conducted an evaluation of the model's hyperparameters, including the F-N aggregation function \( \Psi(\cdot) \) and the fusion function \(\Phi(\cdot)\) for global consensus and local evidence. Detailed results are provided in the \textbf{Supplementary Material}.
\section{Conclusion}
In this study, we introduced the TUNED model to address the challenges of multi-view classification, particularly in scenarios with conflicting views and high uncertainty. Our approach successfully integrates local and global feature-neighborhood structures, significantly enhancing robustness and driving improvements in classification performance. However, a limitation of this work is that it has not yet been tested in scenarios involving incomplete multi-view data, where missing views could potentially degrade the feature-neighborhood structure. Future work will focus on extending our approach to handle incomplete multi-view classification, ensuring the model's robustness even when some views are missing or partially observed.

\bibliography{aaai25}
\newpage
\section{\LARGE Appendix.}
\section{Discussion and analysis}
The proposed Selective Markov Random Field (S-MRF) fusion framework addresses key limitations inherent in traditional Dempster-Shafer (D-S) combination methods, particularly as seen in approaches like ECML~\cite{xu2024reliable}.

\subsection{Challenges with D-S fusion in ECML}

In the ECML framework, it was established that fusing two opinions (\(\omega\)) could be mathematically represented as averaging two pieces of evidence, using the following formula:
\begin{equation}
\boldsymbol{\omega} = \boldsymbol{\omega}^{1} \underline{\lozenge} \boldsymbol{\omega}^{2} \underline{\lozenge} \dots \underline{\lozenge} \boldsymbol{\omega}^{V}.
\end{equation}
This formula is effectively implemented in the code snippet:

\begin{lstlisting}[language=Python]
evidence_a = evidences[0]
for i in range(1, self.num_views):
    evidence_a = (evidences[i] + evidence_a) / 2
\end{lstlisting}

This code implements the D-S combination rule by sequentially averaging the evidence from different views. However, this method has significant drawbacks:

\begin{itemize}
    \item \textbf{Order Sensitivity}: The final fusion result is highly dependent on the order in which evidence is combined. The first piece of evidence significantly biases the outcome, which can lead to unreliable results if the initial evidence is not representative.
    \item \textbf{Vulnerability to Conflict}: The D-S fusion framework does not adequately handle conflicting evidence, leading to a diluted final decision when evidence from different views is contradictory.
\end{itemize}

\subsection{Addressing limitations with S-MRF}

The S-MRF framework resolves these issues by introducing a more sophisticated and dynamic approach:

\begin{itemize}
    \item Unlike the D-S method, which treats all evidence equally, S-MRF selectively integrates evidence based on its relevance and consistency within the global context of the data. This method ensures that the final fusion is not disproportionately influenced by any single view, providing a more balanced and accurate representation.
    \item S-MRF leverages the underlying neighborhood structure to down-weight conflicting or noisy evidence, focusing on the most reliable and consistent information. This selective process allows the model to maintain robustness even when views are in conflict.
\end{itemize}

\subsection{Beyond intricate loss functions}

While earlier EDL-based methods like ECML~\cite{xu2024reliable} and TMDL-OA~\cite{Liu_Yue_Chen_Denoeux_2022} attempted to address these issues by incorporating intricate loss functions and hand-crafted conflict degree measures, these approaches require careful design and hyperparameters tuning, which can be both complex and error-prone.

In contrast, our proposed loss function \(\mathcal{L}_{\text{con}}\) achieves competitive and often superior performance without needing such complex mechanisms. By leveraging the statistical properties of evidence, including cosine similarity, variance, and graph-based regularization, \(\mathcal{L}_{\text{con}}\) provides a straightforward yet powerful way to achieve robust multi-view fusion.

In conclusion, the S-MRF framework, combined with our proposed $\mathcal{L}_{\text{con}}$, offers a significant leap forward compared to traditional EDL-based methods. By providing a more reliable and context-aware fusion process, S-MRF effectively addresses issues like order sensitivity and conflict handling, all while simplifying the integration process without the need for intricate loss functions. This approach consistently delivers superior performance across diverse multi-view datasets. Future refinements to D-S-based methods could benefit from adopting dynamic weighting and advanced conflict resolution strategies, as exemplified by the S-MRF framework, to better balance simplicity and effectiveness in multi-view learning.
\section{Experiment setup}
\subsection{Dataset}
As for \textbf{normal datasets}, \textbf{PIE}\footnote{http://www.cs.cmu.edu/afs/cs/project/PIE/MultiPie/Multi-Pie/Home.html} contains 680 instances belonging to 68 classes. Intensity, LBP, and Gabor features are extracted as three different views.
\textbf{HandWritten}\footnote{https://archive.ics.uci.edu/ml/datasets/Multiple+Features} consists of 2000 instances of handwritten numerals ranging from '0' to '9', with 200 patterns per class. It is represented using six feature sets, each providing a different view of the data. The six views include Fourier coefficients of character shapes (FOU), profile correlations (FAC), Karhunen-Loeve coefficients (KAR), pixel averages, Zernike moments (ZER), and morphological features (MOR).
\textbf{Scene15}\footnote{https://doi.org/10.6084/m97007177.v1} includes 4485 images from 15 indoor and outdoor scene categories. Three types of features are extracted: GIST, PHOG and LBP. 
\textbf{Caltech101}\footnote{http://www.vision.caltech.edu/Image Datasets/Caltech101} comprises 8677 images from 101 classes. The first 10 categories are selected, and two deep features (views) are extracted using the DECAF and VGG19 models.
\textbf{CUB}\footnote{http://www.vision.caltech.edu/visipedia/CUB-200.html}consists of 11788 instances associated with text descriptions of 200 different categories of birds.This study focuses on the first 10 categories, extracting image features using GoogleNet and corresponding text features using doc2vec.
\textbf{Animal}\footnote{https://hannes.nickisch.org/papers/articles/lampert13attributes.pdf}includes 10,158 images distributed across 50 distinct classes,featuring two sets of deep features extracted using DECAF and VGG19.
\textbf{ALOI}\footnote{https://elki-project.github.io/datasets/multi\_view} includes 10,158 images of 1000 small objects,featuring four sets of deep features.
\textbf{NUS-WIDE-OBJECT}\footnote{https://lms.comp.nus.edu.sg/wp-content/uploads/2019/} (NUS) consists of 30,000 images of 31 classes. Each instance is described as 5 views, including Color Histogram, block-wise Color Moments, Color Correlogram, Edge Direction Histogram, and Wavelet Texture. As for \textbf{conflictive datasets}, we constructed them as in~\cite{xu2024reliable}, and each method was run 10 times, with the average and standard deviation reported.

\subsection{Compared methods}
The baselines based on feature fusion include: (1) \textbf{EDL} (Evidential Deep Learning)~\cite{sensoy2018evidential} employs the Dirichlet distribution to explicitly model uncertainty, representing the confidence of predictions. (2) \textbf{DCCAE} (Deep Canonically Correlated AutoEncoders)~\cite{wang2015deep} is a classical method that employs autoencoders to find a common representation across multiple views. (3) \textbf{CPM-Nets} (Cross Partial Multi-view Networks)~\cite{zhang2019cpm} is a state-of-the-art multi-view feature fusion method that learns a versatile representation to handle complex correlations among different views. (4) \textbf{DUA-Nets} (Dynamic Uncertainty-Aware Networks)~\cite{geng2021uncertainty} is an uncertainty-aware method that uses reversal networks to integrate intrinsic information from different views into a unified representation. (5) \textbf{TMC} (Trusted Multi-view Classification)~\cite{han2021trusted} is a pioneering uncertainty-aware method designed to address uncertainty estimation, providing reliable classification results. (6) \textbf{TMDL-OA} (Trusted Multi-View Deep Learning with Opinion Aggregation)~\cite{Liu_Yue_Chen_Denoeux_2022} is a state-of-the-art multi-view decision fusion method based on evidential deep neural networks, introducing a consistency measure loss for trustworthy learning outcomes. (7) \textbf{ECML} (Evidential Conflictive Multi-view Learning)~\cite{xu2024reliable} is a method designed for reliable conflictive multi-view learning, which constructs view-specific opinions and aggregates them using a conflictive opinion strategy to model multi-view common and specific reliabilities. (8) \textbf{RMVC} (Robust Multi-view Classification)~\cite{yue2024evidential} is a robust multi-view classification method that uses an evidential dissonance measure to assess the quality of views under adversarial attacks, enhancing model robustness through dissonance-aware belief integration. 

\subsection{Quantative results}
\begin{table*}[htbp]
\centering
\caption{Classification accuracy (ACC), F1-score, and Area Under the Curve (AUC) for different methods on HandWritten and Scene15 datasets under normal and conflict conditions. \emph{Average} refers to fusion using the arithmetic mean of the evidence, \emph{DST} employs the D-S combination rule for evidence fusion, and \emph{S-MRF} represents the proposed selective Markov Random Field (MRF) method for evidence fusion. All metrics are expressed as percentages (\%).}
\label{tab:ablation_v2}
\small
\setlength{\tabcolsep}{3pt} 
\begin{tabular}{lcccccccccccc}
\toprule
\textbf{Method} & \multicolumn{3}{c}{\textbf{HandWritten}} & \multicolumn{3}{c}{\textbf{HandWritten (Conflict)}} & \multicolumn{3}{c}{\textbf{Scene15}} & \multicolumn{3}{c}{\textbf{Scene15 (Conflict)}} \\
\cmidrule(lr){2-4} \cmidrule(lr){5-7} \cmidrule(lr){8-10} \cmidrule(lr){11-13}
& \textbf{ACC} $\uparrow$ & \textbf{F1-score} $\uparrow$ & \textbf{AUC} $\uparrow$ & \textbf{ACC} $\uparrow$ & \textbf{F1-score} $\uparrow$& \textbf{AUC} $\uparrow$ & \textbf{ACC} $\uparrow$& \textbf{F1-score} $\uparrow$& \textbf{AUC} $\uparrow$& \textbf{ACC} $\uparrow$& \textbf{F1-score} $\uparrow$ & \textbf{AUC} $\uparrow$ \\
\midrule
w/o GCN & 95.75 & 95.71 & 99.93 & 95.75 & 95.70 & 99.88 & 69.23 & 65.87 & 94.54 & 64.21 & 61.66 & 90.06 \\
w/o Consensus Evidence & 96.50 & 96.57 & 99.89 & 94.25 & 94.07 & 99.52 & 70.79 & 68.33 & 94.33 & 64.88 & 61.81 & 91.02 \\
w/o $\mathcal{L}_{\text{con}}$ & 96.50 & 96.45 & 99.88 & 94.50 & 94.44 & 99.59 & 70.79 & 68.11 & 94.32 & 65.44 & 62.52 & 91.28 \\
w/ Average & 96.25 & 96.24 & 99.84 & 96.00 & 95.92 & 99.84 & 70.90 & 68.30 & 94.26 & 65.33 & 62.70 & 91.31 \\
w/ DST & 95.75 & 95.71 & 99.93 & 91.75 & 91.57 & 99.33 & 70.56 & 67.77 & 94.29 & 66.67 & 64.30 & 91.60 \\
Full-model (w/ S-MRF) & \textbf{99.20} & \textbf{99.27} & \textbf{99.98} & \textbf{96.75} & \textbf{96.34} & \textbf{99.92} & \textbf{75.89} & \textbf{74.10} & \textbf{96.45} & \textbf{67.22} & \textbf{65.05} & \textbf{94.72} \\
\bottomrule
\end{tabular}
\end{table*}
\begin{table}
\centering
\caption{Accuracy (\%) of EDL-based methods with varying conflict views on the HandWritten dataset. Methods using the D-S combination rule begin fusion from the first view, making them sensitive to the order of conflicting views.}
\label{tab:robust}
\begin{tabular}{lccc}
\toprule
\textbf{Conflict View(s)} & \textbf{TUNED} & \textbf{ECML} & \textbf{TMDL-OA} \\ 
\midrule
0                   & \textbf{97.5}  & 97.25         & 97.0             \\ 
2                   & 95.75 & 95.75         & \textbf{96.75}            \\ 
4                   & \textbf{94.75} & 79.5          & 82.5             \\ 
0, 2                & \textbf{98.25} & 95.0          & 95.0             \\ 
1, 3                & 92.75 & \textbf{93.5}          & 91.5             \\ 
2, 5                & \textbf{95.0}  & 87.0          & 83.5             \\ 
0, 2, 4             & \textbf{94.5}  & 87.75         & 89.0             \\ 
1, 3, 5             & \textbf{92.0}  & 88.25         & 86.25            \\ 
0, 1, 3, 5          & \textbf{93.0}  & 92.25         & 89.0             \\ 
0, 1, 2, 3, 4       & \textbf{93.25} & 87.75         & 87.75            \\ 
\bottomrule
\end{tabular}
\end{table}
\begin{table}[h]
\centering
\vspace{-2mm}
\caption{Parameter analysis with respect to $\Psi$: Accuracy (\%) comparison of different fusion methods for fusing local feature and neighborhood structure across various normal and conflictive multi-view datasets, including HandWritten, Scene15, and Animal. \emph{SM} refers to the Summation method, \emph{LW} denotes the Linear Weighted average, and \emph{Cross-Attn} represents the Cross-Attention mechanism.}
\label{tab:psi_comp}
\small 
\setlength{\tabcolsep}{4pt} 
\begin{tabular}{lcccccc}
\toprule
\textbf{Method} & \multicolumn{2}{c}{\textbf{HandWritten}} & \multicolumn{2}{c}{\textbf{Scene15}} & \multicolumn{2}{c}{\textbf{Animal}} \\
\cmidrule(lr){2-3} \cmidrule(lr){4-5} \cmidrule(lr){6-7}
 & \textbf{Normal} & \textbf{Conflict} & \textbf{Normal} & \textbf{Conflict} & \textbf{Normal} & \textbf{Conflict} \\
\midrule
SM & \textbf{98.58} & 97.83 & \textbf{71.76} & \textbf{65.37} & 72.18 &\textbf{ 74.93} \\
LW & 95.83 & 95.58 & 65.11 & 58.68 & \textbf{84.35} & 64.16 \\
Cross-Attn & 98.25 & \textbf{98.08} & 71.65 & 64.29 & 84.04 & 64.34 \\
\bottomrule
\end{tabular}
\vspace{-1mm}
\end{table}

\begin{table}[h]
\centering
\vspace{-2mm}
\caption{Parameter analysis with respect to $\Phi$: Accuracy (\%) comparison of different fusion methods for global consensus and local view-specific evidence across various normal and conflicting multi-view datasets, including HandWritten, Scene15, and Animal. \emph{SM} refers to the Summation method, \emph{LW} denotes the Linear Weighted average, and \emph{Cross-Attn} represents the Cross-Attention mechanism.}
\label{tab:phi_comp}
\small 
\setlength{\tabcolsep}{4pt} 
\begin{tabular}{lcccccc}
\toprule
\textbf{Method} & \multicolumn{2}{c}{\textbf{HandWritten}} & \multicolumn{2}{c}{\textbf{Scene15}} & \multicolumn{2}{c}{\textbf{Animal}} \\
\cmidrule(lr){2-3} \cmidrule(lr){4-5} \cmidrule(lr){6-7}
 & \textbf{Normal} & \textbf{Conflict} & \textbf{Normal} & \textbf{Conflict} & \textbf{Normal} & \textbf{Conflict} \\
\midrule
SM & \textbf{98.50} & \textbf{98.00} & 71.63 & \textbf{64.60} & 80.07 & 68.90 \\
LW & 98.00 & \textbf{98.00} & \textbf{73.80} & 64.21 & 88.51 & 71.09 \\
Cross-Attn & 98.00 & 97.88 & 64.44 & 52.68 & \textbf{88.78} &\textbf{ 71.11} \\
\bottomrule
\end{tabular}
\end{table}
\subsubsection{Robust to conflict.}
Table~\ref{tab:robust} evaluates the robustness of different methods against varying conflict view combinations on the HandWritten dataset. Methods based on the D-S combination rule show a clear sensitivity to the order and combination of conflicting views due to their reliance on the sequence of evidence fusion. This inherent order sensitivity leads to disproportionate dependence on certain views, which negatively impacts their robustness when faced with conflicts. The results demonstrate that these methods perform inconsistently across different conflict scenarios, highlighting their vulnerability to variations in conflict view order.

In contrast, our proposed method exhibits greater robustness against conflicts. By avoiding the order sensitivity inherent in D-S-based approaches, our method effectively mitigates the impact of conflicting views, leading to more stable and reliable performance across all scenarios. This indicates that the proposed method not only provides a more balanced integration of evidence but also offers superior resilience to the challenges posed by conflicting data, making it a more reliable choice for complex multi-view tasks.
\subsubsection{Ablation study.}
In our Neighborhood-aware EDL loss, the aggregation function \(\Psi(\cdot)\) fuses the local feature structure \(\mathbf{h^v_n}\) and neighborhood structure \(\mathbf{q^v_n}\). We implement \(\Psi(\cdot)\) in three different ways:
\begin{itemize}
    \item \textbf{Summation}: The simplest approach directly adds the local feature \(\mathbf{h^v_n}\) and neighborhood structure \(\mathbf{q^v_n}\):
    \[
    \Psi(\mathbf{h^v_n}, \mathbf{q^v_n}) = \mathbf{h^v_n} + \mathbf{q^v_n}.
    \]
    
    \item \textbf{Linear Weighted Average}: This method combines \(\mathbf{h^v_n}\) and \(\mathbf{q^v_n}\) through a linear weighted average, where \(\lambda_1\) and \(\lambda_2\) are learnable parameters satisfying \(\lambda_1 + \lambda_2 = 1\):
    \[
    \Psi(\mathbf{h^v_n}, \mathbf{q^v_n}) = \lambda_1 \mathbf{h^v_n} + \lambda_2 \mathbf{q^v_n}.
    \]
    
    \item \textbf{Cross-Attention}: We apply a cross-attention mechanism where \(\mathbf{h^v_n}\) serves as the Query, and \(\mathbf{q^v_n}\) acts as both the Key and Value. The fusion is computed as:
    \[
    \Psi(\mathbf{h^v_n}, \mathbf{q^v_n}) = \text{softmax}\left(\frac{\mathbf{h^v_n} (\mathbf{q^v_n})^{\top}}{\sqrt{d_k}}\right) \mathbf{q^v_n}.
    \]
\end{itemize}
Each method is tested and its performance is shown in the Table~\ref{tab:phi_comp}.

Similarly, the fusion function $\Phi(\cdot)$ in our framework integrates global consensus evidence and local view-specific evidence using three distinct methods: Summation, Linear Weighted Average, and Self-Attention. Each method is tested and its performance is shown in the Table~\ref{tab:phi_comp}. They are formulated respectively as follows.
\begin{itemize}
    \item \textbf{Summation:} The simplest approach directly adds the global consensus evidence $\mathbf{e}^{\diamond}_n$ and the local view-specific evidence $\mathbf{e}^v_n$:
    \[
    \Phi(\mathbf{e}^{\diamond}_n, \mathbf{e}^v_n) = \mathbf{e}^{\diamond}_n + \mathbf{e}^v_n
    \]
    
    \item \textbf{Linear Weighted Average:} This method combines the global and local evidence through a linear weighted average, where $\lambda_1$ and $\lambda_2$ are learnable parameters that satisfy the constraint $\lambda_1 + \lambda_2 = 1$:
    \[
    \Phi(\mathbf{e}^{\diamond}_n, \mathbf{e}^v_n) = \lambda_1 \mathbf{e}^{\diamond}_n + \lambda_2 \mathbf{e}^v_n
    \]
    
    \item \textbf{Cross-Attention:} In this approach, we apply a cross-attention mechanism where the local evidence $\mathbf{e}^v_n$ serves as the Query, and the global consensus evidence $\mathbf{e}^{\diamond}_n$ acts as both the Key and Value. The fusion is computed as:
    \[
    \Phi(\mathbf{e}^{\diamond}_n, \mathbf{e}^v_n) = \text{softmax}\left(\frac{\mathbf{Q} \mathbf{K}^{\top}}{\sqrt{d_k}}\right) \mathbf{V}
    \]
    Here, $\mathbf{Q} = W_Q \mathbf{e}^v_n$ represents the Query vector derived from the local evidence, while $\mathbf{K} = W_K \mathbf{e}^{\diamond}_n$ and $\mathbf{V} = W_V \mathbf{e}^{\diamond}_n$ are the Key and Value vectors derived from the global evidence, respectively. The term $d_k$ denotes the dimensionality of the Key vectors, ensuring appropriate scaling in the attention mechanism.
\end{itemize}
Each of these fusion methods is designed to effectively balance and integrate the complementary information provided by global consensus and local view-specific evidence, enhancing the model's ability to adapt to diverse scenarios.

\subsection{Qualitative results}
\begin{figure}
    \centering
    \begin{minipage}[b]{\linewidth}
        \centering
        \includegraphics[width=\linewidth]{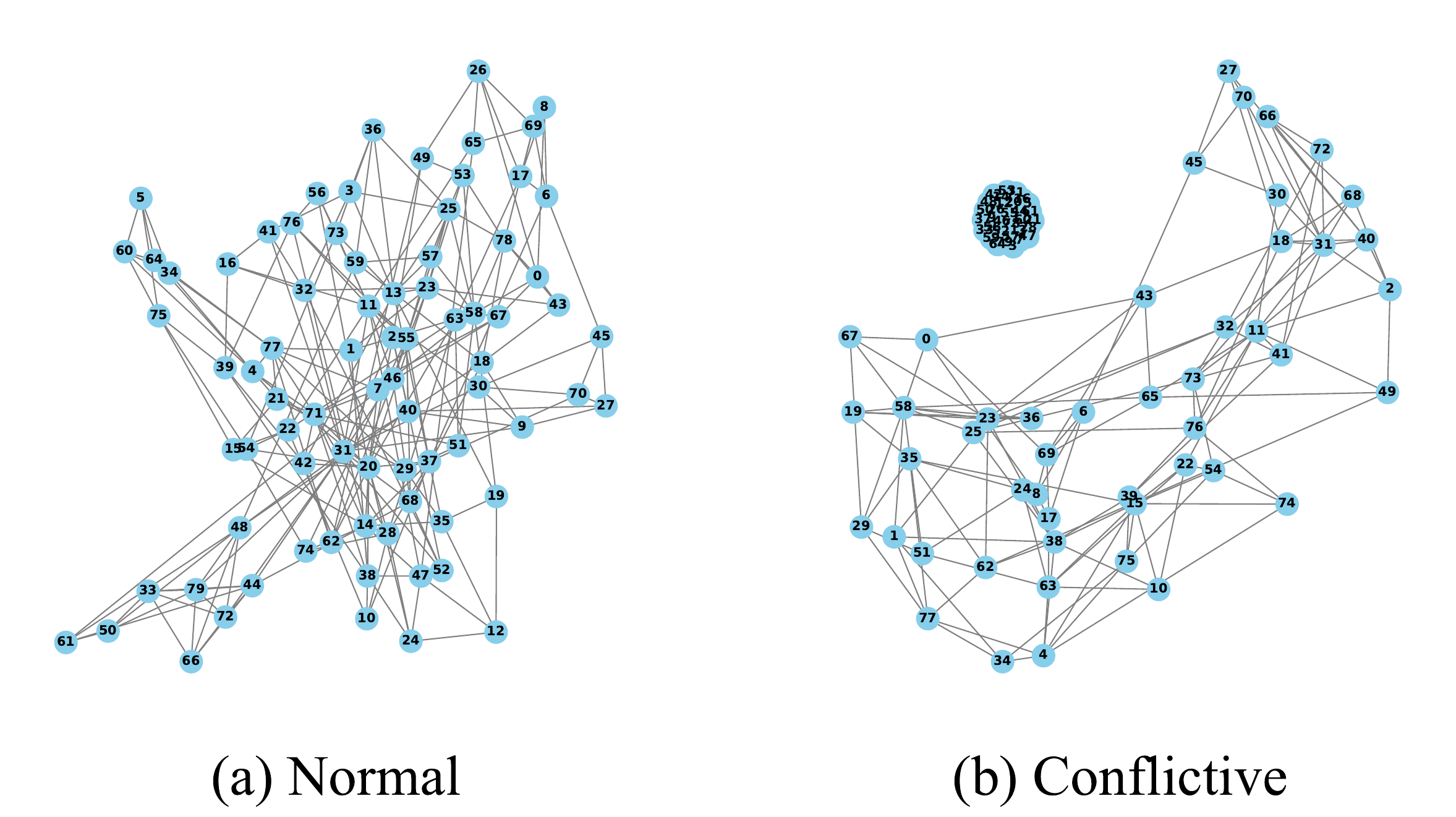}
        \caption*{\romannumeral 1. HandWritten}
    \end{minipage}
    \vspace{2mm}
    \begin{minipage}[b]{\linewidth}
        \centering
        \includegraphics[width=\linewidth]{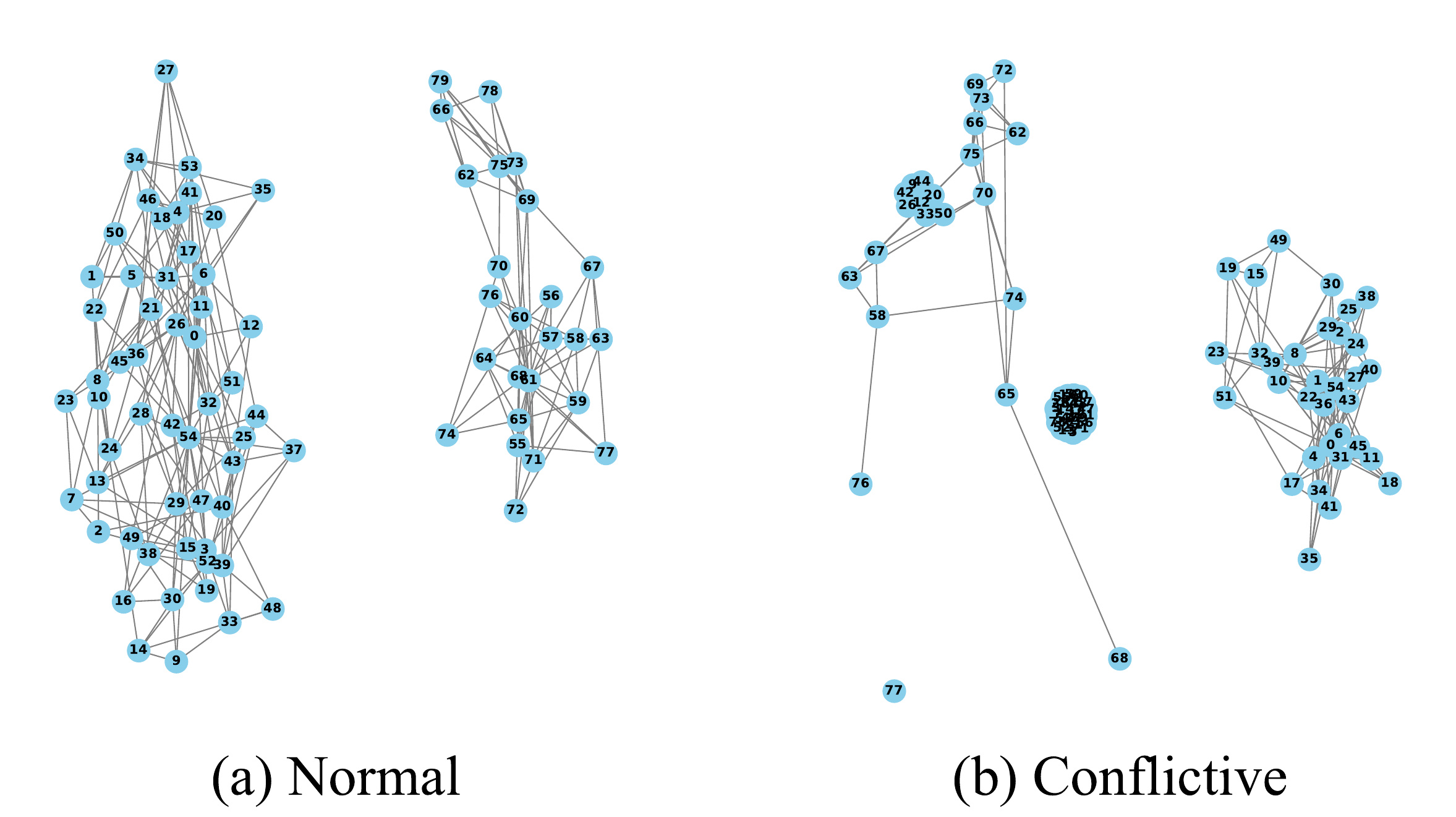} 
        \caption*{\romannumeral 2. Caltech101}
    \end{minipage}
    \vspace{2mm}
    \begin{minipage}[b]{\linewidth}
        \centering
        \includegraphics[width=\linewidth]{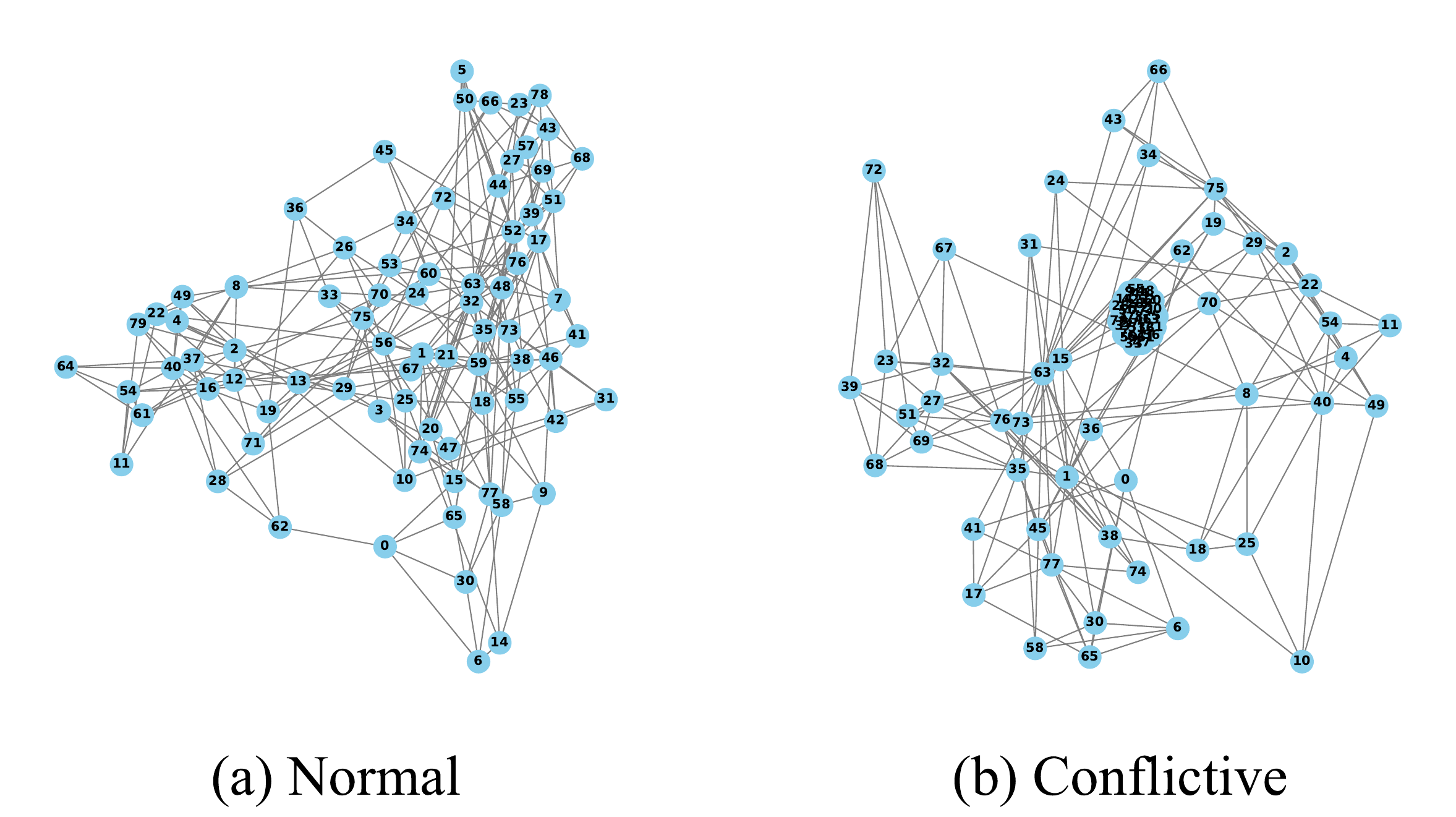} 
        \caption*{\romannumeral 3. Animal}
    \end{minipage}
    \vspace{-2mm}
    \caption{Visualization of neighborhood structures extracted from normal (a) and conflictive (b) multi-view datasets. The adjacency matrix visualizations highlight the differences in connectivity patterns and structural organization between normal and conflict states across three datasets.} 
    \label{fig:graph_vis}
    \vspace{-3mm}
\end{figure}

\subsubsection{Neighborhood struture visualization.} Figure~\ref{fig:graph_vis} showcases the effectiveness of our proposed method in distinguishing between normal and conflictive multi-view datasets. In the normal state (a), the neighborhood structure is densely connected, reflecting consistent and coherent relationships across views. In contrast, the conflictive state (b) shows a fragmented and sparsely connected structure, with noticeable isolated clusters. This clear differentiation underscores our model's ability to accurately capture and represent the underlying structural differences in multi-view data, effectively identifying and separating conflictive instances from normal ones, which is crucial for improving model robustness and performance in complex multi-view scenarios.

\begin{figure}[h]
    \centering
    \includegraphics[width=\linewidth]{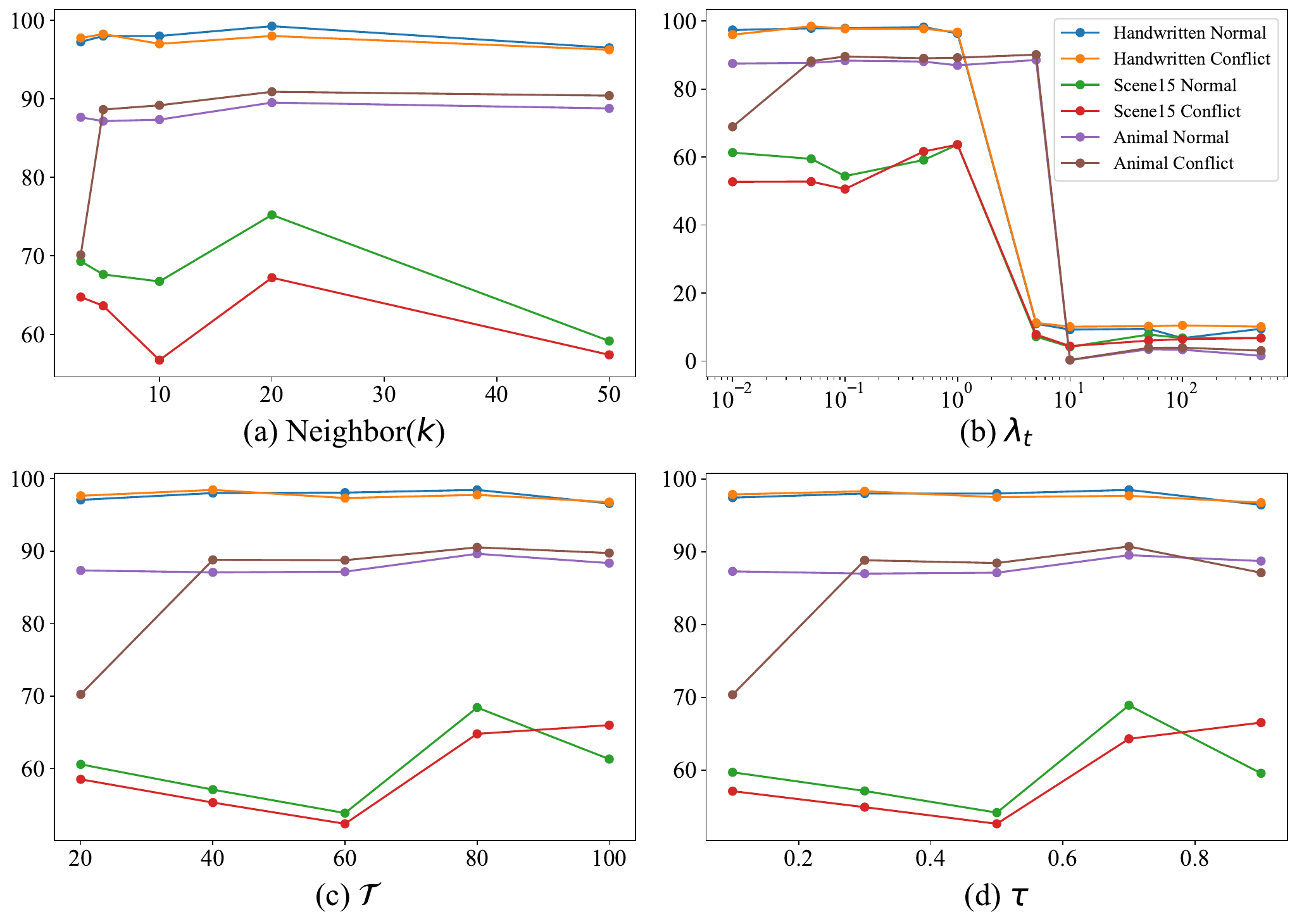}
    \vspace{-2mm}
    \caption{Sensitivity analysis of model performance across various datasets with respect to key parameters under normal and conflictive conditions.} 
    \label{fig:para_analysis}
    \vspace{-3mm}
\end{figure}
\begin{figure}[h]
    \centering
    \includegraphics[width=\linewidth]{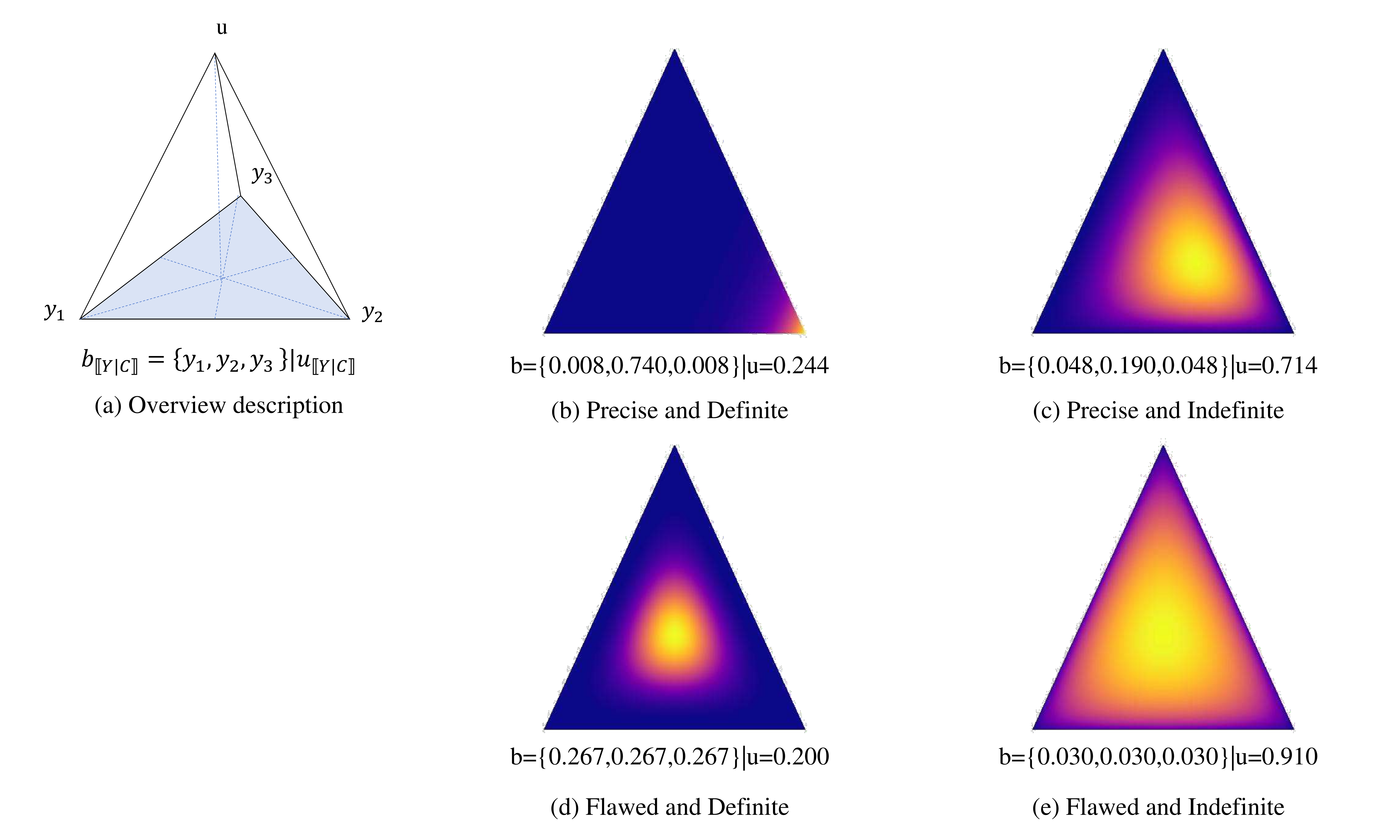}
    \vspace{-2mm}
    \caption{(a) Overview of uncertainty estimation for Dirichlet distributions. (b-e) Typical patterns of Dirichlet distribution for an example 3-class classification task.} 
    \label{fig:dirichlet}
    \vspace{-3mm}
\end{figure}

\subsubsection{Parameter analysis.} Figure~\ref{fig:para_analysis} highlights the sensitivity of the proposed model to key parameters across different datasets. The results demonstrate that while the model maintains robust performance within certain parameter ranges, it exhibits significant sensitivity to variations, particularly under conflictive conditions. This underscores the importance of careful parameter tuning to optimize model performance across diverse and challenging datasets.

\subsubsection{Understanding of opinions.} As shown in the Figure~\ref{fig:dirichlet}, the varying patterns reflect different levels of certainty and uncertainty, enabling clear insight into model confidence and its evidential support. This visualization underscores the model's ability to transparently communicate both predictions and their associated uncertainties, enhancing trust in decision-making.
\end{document}